\documentclass{article}
\usepackage{float}
\usepackage[a4paper,margin=1in]{geometry}

\usepackage{amsmath,amssymb,amsfonts}
\usepackage{bm}

\usepackage{graphicx}
\usepackage{caption}
\usepackage{subcaption}
\usepackage{booktabs}
\usepackage{tabularx}

\usepackage{algorithmic}

\usepackage{cite}
\usepackage{xcolor}
\usepackage{textcomp}

\usepackage{authblk}
\usepackage{setspace}
\usepackage[hidelinks]{hyperref}
\title{Multimodal Deep Learning for Dynamic and Static Neuroimaging:
Integrating MRI and fMRI for Alzheimer Disease Analysis}

\author{Anima Kujur}
\author{Zahra Monfared}

\affil[]{Interdisciplinary Centre for Scientific Computing (IWR),
Department of Mathematics and Computer Science,
Heidelberg University, Germany\\ \texttt{\{anima.kujur, zahra.monfared\}@iwr.uni-heidelberg.de}}

\date{}
\begin{document}
\maketitle


\begin{abstract}
Magnetic Resonance Imaging (MRI) provides detailed structural information, while functional MRI (fMRI) captures temporal brain activity. In this work, we present a multimodal deep learning framework that integrates MRI and fMRI for multi-class classification of Alzheimer’s Disease (AD), Mild Cognitive Impairment, and Normal Cognitive State. Structural features are extracted from MRI using 3D convolutional neural networks, while temporal features are learned from fMRI sequences using recurrent architectures. These representations are fused to enable joint spatial-temporal learning. Experiments were conducted on a small paired MRI–fMRI dataset (29 subjects), both with and without data augmentation. Results show that data augmentation substantially improves classification stability and generalization, particularly for the multimodal 3DCNN-LSTM model. In contrast, augmentation was found to be ineffective for a large-scale single-modality MRI dataset. These findings highlight the importance of dataset size and modality when designing augmentation strategies for neuroimaging-based AD classification.
\end{abstract}

\section{Introduction}
\label{sec:introduction}
Alzheimer’s Disease (AD) \cite{AD-review-Mirabian, Liu-Alzheimer’s-Disease}, a progressive neurodegenerative disorder\cite{AD-LSTM,AD-segmentation-Coupé}, is characterized by profound structural and functional changes in the brain. These alterations manifest as atrophy in key regions such as the hippocampus and cortical areas, as well as disturbances in intrinsic brain networks. Magnetic Resonance Imaging (MRI)  serves as a vital clinical tool for capturing the static, high-resolution structural details of the brain, thereby allowing clinicians and researchers to investigate cortical thinning, volumetric changes, and lesion formation \cite{AD-multimodal-MRI}. On the other hand, functional MRI (fMRI) offers insight into dynamic brain activity by tracking fluctuations in Blood-Oxygen-Level-Dependent (BOLD) signals \cite{Multimodal}. Analyzing these signals can uncover functional connectivity patterns and identify regions of the brain whose activities become disrupted in early or advanced stages of AD.

Despite the complementary nature of MRI and fMRI, these modalities are often studied in isolation \cite{Li-fMRI-study}. Standard practice, for instance, involves applying  classification\cite{CNN–RNN–LSTM} techniques on MRI scans to detect structural abnormalities \cite{func-conn}, while employing separate methods on fMRI data to characterize the temporal dynamics \cite{dyna-adj} of brain networks \cite{fmriDynamic-Xinyue-Yan}.

%
However, such a separated approach may fail to capture the complex relationship between structural lesions and functional changes that characterize AD classification \cite{Sun-mri-AD}.
Recent advances in 
Artificial Intelligence (AI)\cite{AD-Amir}—and particularly Deep Learning (DL)\cite{Liu-gereralisable-DL}—provide a powerful set of tools to integrate these heterogeneous data sources into a single unified framework, thereby enhancing diagnostic accuracy and prognostic analyses\cite{MCC-F1-score}.

In this work, we propose a multimodal DL framework designed to leverage both the spatial information from MRI scans and the temporal 
dynamics from fMRI data for comprehensive AD analysis. Building upon Convolutional Neural Networks (CNN)\cite{Guo-CNN-RNN}, Recurrent Neural Networks (RNN)\cite{AD-LSTM}, and generative models\cite{transfer-learning}, our approach addresses several challenges: (i) efficiently extracting robust structural features from MRI, (ii) capturing dynamic functional patterns from fMRI time-series, and (iii) fusing these high-dimensional representations into an integrated pipeline. By jointly learning from both modalities, our model aims to capture nuanced correlations between anatomical degeneration and functional disruptions. Furthermore, we incorporate generative techniques to synthesize missing data or augment existing datasets, thereby mitigating common challenges such as limited sample size and missing data points. 
%
The proposed framework offers several practical applications. 
First, it can serve as a powerful diagnostic tool, providing clinicians with both structural markers (e.g., tissue atrophy) and functional indicators (e.g., network connectivity changes). Second, it facilitates patient monitoring by tracking how the interplay between structural and functional changes evolves over time, 
allowing for more timely interventions.
%
Third, its ability to generate synthetic data from one modality to another can aid research in settings where data is limited, thereby extending the utility of DL models in clinical environments. 
%
%
Overall, these advancements could significantly enhance both our fundamental understanding of AD and the development of more personalized treatment strategies.

This paper is organized as follows. Section \ref{sec:RW} discusses related work in multimodal neuroimaging and DL. Section \ref{sec:MM} provides an overview of our data acquisition and preprocessing pipeline, detailing 
our methods for handling both MRI and fMRI data. Further,  it describes our proposed DL architecture, highlighting the spatial-temporal fusion strategies. Section \ref{sec:ResultsD} presents our experimental results, while Section \ref{sec:Dis} discusses the implications of our findings and potential avenues for future research. Finally, Section \ref{sec:Con} concludes the paper.

\section{Related Work}
\label{sec:RW}

Recent advancements in neuroimaging\cite{Liu-Alzheimer’s-Disease} have leveraged DL\cite{Liu-gereralisable-DL} techniques to improve the diagnosis and monitoring of 
AD
\cite{Liu-Alzheimer’s-Disease}. Traditional machine learning\cite{AUC-decision} methods, such as Support Vector Machines (SVM)\cite{svm-ad} and Random Forests\cite{random-forest} \cite{RandomFAD}\cite{random}, have been employed to classify MRI and functional MRI data separately. However, these approaches often fail to fully capture the complex spatial-temporal\cite{AD-spatio-temporal-Nobukawa}\cite{AD-Wang-Spatial-Temporal-Dependency} relationships between anatomical structures and functional activity patterns.\\
Multimodal DL\cite{AD-multimodal-MRI} frameworks have emerged as a promising solution for integrating structural and functional neuroimaging data. CNNs\cite{Guo-CNN-RNN} have been widely used for  MRI analysis, while RNNs\cite{rnn} 
especially LSTM networks \cite{Multi-Modal-3D-CNN-RNN} have been applied to fMRI time-series data \cite{Resting-state-Tang} \cite{multimodal-fusion}. Recent studies have explored the combination of CNNs and RNNs for multimodal fusion\cite{Multi-Modal-3D-CNN-RNN}\cite{Multimodal}, improving classification performance by effectively capturing both spatial and temporal dependencies. Resting-state fMRI studies reveal altered connectivity patterns in 
AD beyond the Default Mode Network (DMN). Agosta et al. \cite{Agosta} found decreased DMN and increased frontal network connectivity in AD, while Meskaldji et al.\cite{Meskaldji-longterm-memory} linked functional discordance to memory performance in Mild Cognitive Impairment (MCI). 
Khazaee et al.\cite{Khazaee} combined graph theory and machine learning, achieving a high classification accuracy for AD detection.

Several studies have integrated resting-state fMRI and MRI for 
AD
classification. Hojjati et al. \cite{Hojjati-StructuralMRI-Resting-StatefMRI} used graph theory and SVM to distinguish MCI converters from non-converters.  Ramzan et al. \cite{Ramzan-Multi-Class-Classification} leveraged Residual Neural Network (ResNet-18) with transfer learning on resting state fMRI (rs-fMRI), while Ghafoori \cite{Ghafoori-rs-fMRI} and Shalbaf combined clinical data and rs-fMRI using 3DCNN for MCI-to-AD conversion prediction. 
Recent studies have explored deep DL and graph-based approaches for 
AD
classification using rs-fMRI. Khazaee et al.\cite{Khazaee-Ali-class} applied graph-theoretical measures and machine learning to identify alterations in brain networks. Ahmadi et al.\cite{Ahmadi} introduced sparse graph functional connectivity analysis to refine correlation matrices, improving the distinction between AD and healthy subjects. Tuovinen et al.\cite{Tuovinen-fMRI} investigated the variability of
the BOLD
signal in Alzheimer’s disease and found a significant increase in BOLD signal fluctuations at cardiorespiratory frequencies. Using data from three independent cohorts, they demonstrated that this increased variability serves as a robust biomarker for distinguishing AD patients from healthy controls. Their findings suggest that abnormal cerebral perfusion and cerebrospinal fluid convection observed in AD may be linked to impaired glymphatic clearance, offering new insights into disease mechanisms and potential early detection strategies.

Recent research has explored DL and imaging techniques for AD diagnosis and classification. Liu et al. \cite{Liu-CSEPC} proposed Cross-Scale Equilibrium Pyramid coupling (CSEPC), a multimodal DL framework for classifying AD using small-sample MRI and fMRI data, achieving high accuracy and AUC scores. 
Li et al. \cite{Li-fMRI-study} investigated task-fMRI patterns in subgroups of MCI patients within Traditional Chinese Medicine (TCM), which demonstrated distinct differences in neural activation.

Mirabian et al. \cite{AD-review-Mirabian} reviewed AI applications in differentiating AD and Frontotemporal Dementia (FTD), highlighting the effectiveness of SVM and ResNet models. Sun et al. \cite{Sun-mri-AD} conducted a bibliometric analysis of MRI-based AD research, identifying trends in DL applications. Tang et al. \cite{Resting-state-Tang} examined Cerebrovascular Reactivity (CVR) changes in the precuneus using resting-state fMRI, demonstrating its potential as a biomarker for cognitive decline. 
These studies highlight
the growing role of AI and neuroimaging in AD research. Li et al.\cite{Li-fMRI-study} used a Multiscale Neural Model Inversion (MNMI) framework on fMRI data to map Excitation-Inhibition (E-I) imbalance in AD which identifies disrupted neural circuits in limbic and cingulate regions. Wang \cite{AD-Wang-Spatial-Temporal-Dependency}  introduced a ViT-based transfer learning model for fMRI, improving cognitive decline classification. Jahani et al.\cite{T1-MRI-Jahani} applied DL on FDG-PET and MRI, showing FDG-PET as a more effective biomarker for MCI and AD. Yan et al.\cite{fmriDynamic-Xinyue-Yan} developed a Multi-View Topology Assisted Dynamic Graph Learning (MTDGL) framework, integrating spatiotemporal and topological features for AD detection. These studies advance AI-driven neuroimaging for AD diagnosis.

Despite these advancements, challenges remain in effectively fusing high-dimensional data from multiple modalities. 
Many existing methods treat structural and functional information as independent features rather than 
their inherent interdependence.
%
 In this work, we address this limitation by introducing a DL architecture that performs spatial-temporal integration, enabling more accurate diagnosis and improved monitoring of disease classification.

\section{Materials and Methods}
\label{sec:MM}
\subsection{Data}
\begin{figure}[ht]\centering\includegraphics[width=0.8\linewidth]{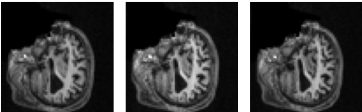} \\[\baselineskip]\includegraphics[width=0.8\linewidth]{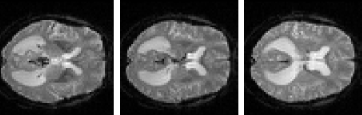} \\[\baselineskip]\includegraphics[width=0.8\linewidth]{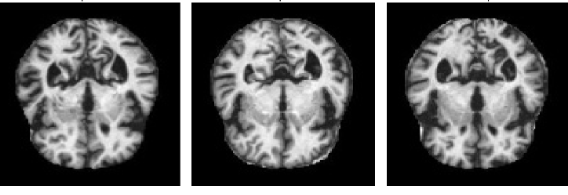} \\[\baselineskip]\caption{Row 1: MRI images (HD), Row 2: fMRI images (HD), Row 3: MRI images (Kaggle) } 
\label{fig:sample images}
\end{figure}

The dataset used in this study consists of multimodal neuroimaging data, including MRI and fMRI scans, collected for AD classification analysis. Two datasets have been used for the implementation of the proposed method. First dataset has been collected from Havard Dataverse (HD)\cite{haverd-data}. This data is categorized into three diagnostic groups: AD, Mild Cognitive Impairment (MCI), and Normal Cognitive State (NCS), with equal representation of subjects across categories. Each subject's folder contains both MRI and fMRI files in the Neuroimaging Informatics Technology Initiative (NIfTI) format, labeled as 'structural' or 'functional'. Fig.~\ref{fig:sample images} represents the sample images from implemented datasets. The MRI data provides high-resolution anatomical details, while the fMRI data captures temporal brain activity. The dataset includes a total of 29 subjects, each MRI image file is a grayscale image with dimension (224$\times$256$\times$176). The fMRI image files have  216 temporal frames of dimension (64$\times$64$\times$32). The Second dataset is collected from Kaggle (https://www.kaggle.com/datasets) to validate the proposed method. The dataset consists of preprocessed 2D MRI images divided into Train (8192), Val (2048), and Test (1279) having four categories which includes: very mild impairment, no impairment, moderate impairment, and mild impairment. The Kaggle dataset was not used for cross-dataset validation of the multimodal MRI–fMRI framework due to differences in class definitions and the absence of paired fMRI data. Instead, it was used exclusively to analyze the effect of data augmentation on large-scale single-modality MRI classification, serving as a complementary study to contextualize augmentation behavior under different data regimes.
\subsubsection{Data Preprocessing}
The dataset consists of MRI  and fMRI  scans categorized into three groups: AD, MCI , and NCS. Preprocessing steps include skull stripping, motion correction, spatial normalization, and intensity standardization.
\subsubsection{Feature Extraction}
To extract meaningful features from the neuroimaging data, we employ a CNN-based encoder for MRI and an LSTM-based model for fMRI. The CNN extracts spatial features, while the LSTM captures dynamic temporal patterns in the fMRI signals. The extracted features are then normalized and concatenated to form a unified multimodal representation.
%
%
The MRI data has been preprocessed and reshaped into a 5D tensor of shape (29, 224, 256, 176, 1), where 29 is the number of subjects, followed by the 3D volume dimensions, and a final channel dimension. Similarly, the functional MRI data has been processed into shape (29, 216, 64, 64, 32, 1), representing 29 subjects, each with 216 time frames of 3D fMRI volumes. Each subject is labeled according to their clinical diagnosis: AD (2.0), MCI (1.0), or NCS (0.0). These labels are stored in a one dimensional array of shape $(29, )$.
The structural data captures critical anatomical features, and the functional data provides temporal information on brain activity, making this dataset suitable for exploring the interplay between brain structure and function in AD classification. All data was normalized and preprocessed to ensure uniformity and consistency, enabling effective training of the proposed DL model.

\subsection{Data Augmentation}

Data augmentation is an essential step in DL pipelines for medical image analysis, particularly in neuroimaging, where annotated datasets are typically small due to acquisition cost and expert labeling requirements. In our study, we designed a comprehensive augmentation strategy to improve the generalizability and robustness of our multimodal model by synthetically enlarging the training dataset through diverse transformations. Initially, the dataset contained $29$ paired samples, covering three classes: AD, MCI, and NCS. The initial training and validation class distribution was moderately balanced with 7 AD, 8 MCI, and 8 NCS in training, and 2 each in validation. To improve data diversity, we implemented domain-specific augmentation functions tailored separately for MRI and fMRI modalities. The function \texttt{resize\_mri\_volume} standardizes MRI input volumes to a fixed spatial resolution of $128 \times 128 \times 176$, ensuring dimensional uniformity across samples. Spatial variability was introduced using \texttt{random\_rotate\_3d} and \texttt{random\_shift\_3d}, which apply random affine transformations—specifically rotation around three anatomical axes and spatial translations within a controlled range—thereby emulating inter-subject and scanner-induced variability. Furthermore, \texttt{random\_intensity\_scale} perturbs voxel intensity distributions by multiplicative scaling factors sampled from a uniform range, simulating scanner contrast variability. To simulate acquisition noise, \texttt{add\_gaussian\_noise} adds random Gaussian perturbations to the image intensities. These transformations make the model invariant to typical variations encountered in clinical scans.

For fMRI data, the function \texttt{augment\_fmri} incorporates both spatial and temporal distortions. Apart from affine transformations, this function introduces temporal shifts, mimicking inter-scan misalignments, and synthetic motion artifacts, thus providing temporal deformation realism in dynamic sequences. This augmentation was especially critical for improving LSTM learning across time steps by encouraging the model to generalize to temporally misaligned signals. Post augmentation, the dataset was expanded to 319 samples. The dataset was then stratified into three splits: training $(223 \text{ samples})$, validation $(64 \text{ samples})$, and testing $(32 \text{ samples})$, maintaining class distribution in each split. This augmented dataset not only addresses the overfitting risk associated with limited data but also strengthens model resilience under unseen acquisition conditions. By simulating real-world variabilities, our augmentation pipeline significantly enhanced the reliability and performance of the DL model for AD classification.

\subsection{Methodology}
\begin{figure}[htbp]
    \centering
    \includegraphics[width=0.25\textwidth]{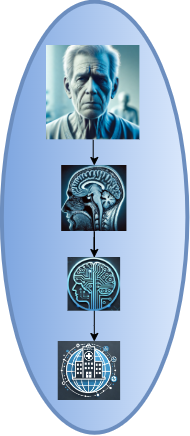} \hfill
    \includegraphics[width=0.49\textwidth]{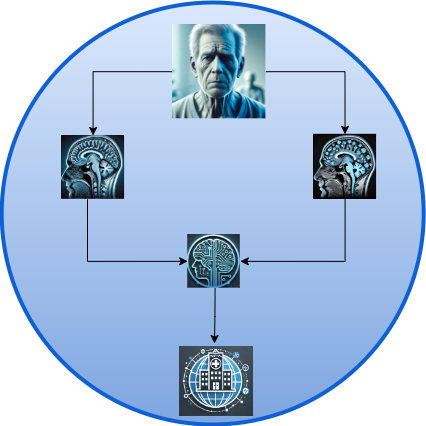}
    \caption{(a) Single Modal. (b) Multi-Modal.}
    \label{fig:single-multi modal}
\end{figure}

Conventional methods employ single modal approach for AD analysis as shown in the Fig.~\ref{fig:single-multi modal}(a).  The proposed methodology uses multimodal neuroimaging data, specifically MRI and fMRI, to detect the AD. Fig.~\ref{fig:single-multi modal}(b) represents the multi-modal approach. The workflow begins with the preprocessing of MRI and fMRI data to ensure uniform dimensions and scaling. MRI data is processed through a CNN to extract spatial features capturing anatomical details. Similarly, fMRI data, represented as a time-series of slices, is passed through a slice-wise CNN to extract spatial features for each temporal frame. The extracted spatial features from both modalities are then fused to form a comprehensive latent feature space, representing both structural and functional aspects of the brain.
 
\begin{figure}[ht]
\centering
\begin{minipage}[t]{0.40\textwidth}
    \centering
    \includegraphics[width=\linewidth]{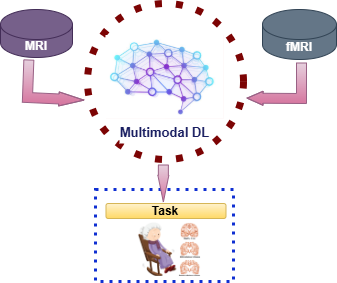}
    \captionof{figure}{Proposed Framework}
    \label{fig:proposed idea}
\end{minipage}
\hfill
\begin{minipage}[t]{0.48\textwidth}
    \centering
    \includegraphics[width=\linewidth]{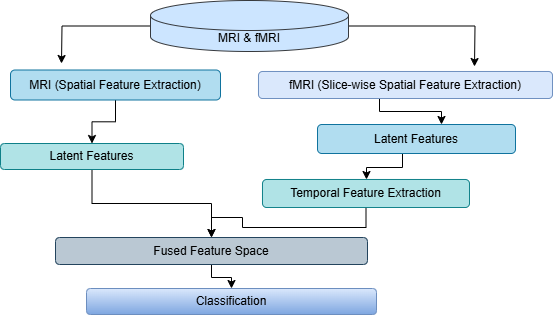}
    \captionof{figure}{Implementation Pipeline of the Proposed Approach}
    \label{fig:flow diagram}
\end{minipage}
\end{figure}

To model temporal dependencies in the fMRI data, the fused latent features are passed through LSTM network. The LSTM captures temporal patterns and dynamic interactions across the time-series data. These extracted temporal features are then utilized for classification or regression, depending on the task, to detect AD. The methodology integrates the spatial features of MRI and fMRI with the temporal dynamics captured from the LSTM, creating a unified framework for comprehensive neuroimaging analysis as shown in the proposed method in Fig.~\ref{fig:proposed idea}. This approach aims to bridge structural and functional modalities for robust and interpretable disease classification modeling. Fig.~\ref{fig:flow diagram} shows the implementation pipeline for the proposed framework.

\subsection{DL Architectures}
\begin{figure}[ht]

    \centering
    \includegraphics[width=0.48\textwidth]{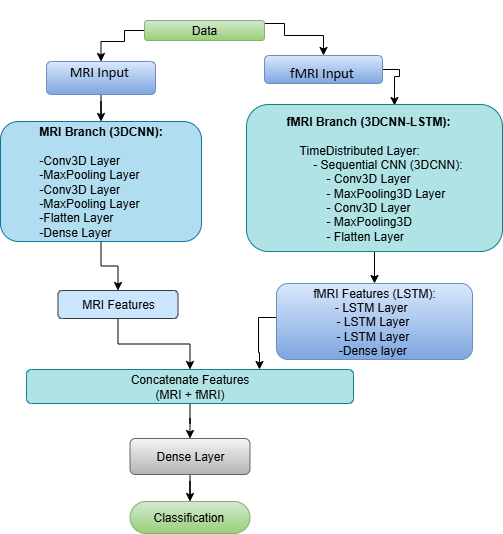} \hfill
    \includegraphics[width=0.48\textwidth]{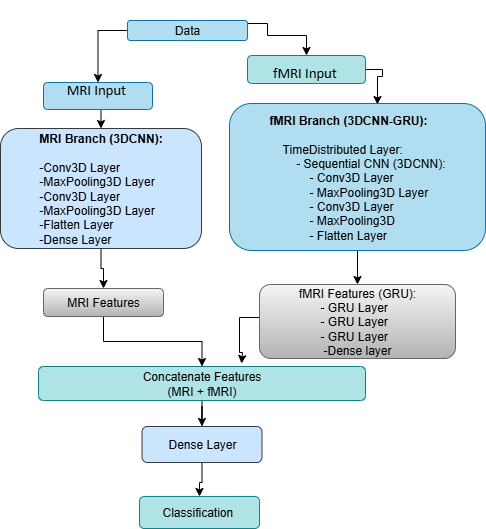}
    \caption{Multi-Modal Architectures: (a) 3DCNN-LSTM Model (b) 3DCNN-GRU Model}
    \label{fig:LSTM GRU}
\end{figure}

\begin{figure}[h]\centering\includegraphics[width=0.48\textwidth]{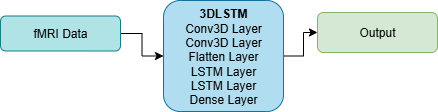} \\[\baselineskip]\includegraphics[width=0.48\linewidth]{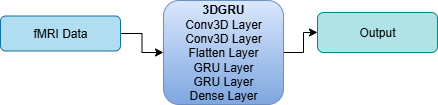} \\[\baselineskip]\includegraphics[width=0.48\linewidth]{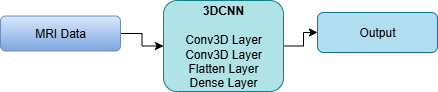} \\[\baselineskip]\caption{Single Modal Architectures: Row 1: 3DLSTM, Row 2: 3DGRU, Row 3: 3DCNN}
\label{fig:single modals}
\end{figure}

To ensure effective learning from multimodal neuroimaging data, we implemented and trained several DL architectures, including 3DCNN-LSTM (Fig.~\ref{fig:LSTM GRU}a), 3DCNN-GRU (Fig.~\ref{fig:LSTM GRU}b), 3DLSTM (Fig.~\ref{fig:single modals}, Row 1), 3DGRU (Fig.~\ref{fig:single modals}, Row 2), and 3DCNN (Fig.~\ref{fig:single modals}, Row 3) models. Each of these architectures was designed to capture both spatial and temporal patterns in MRI and fMRI data. The model architectures clearly present the number of layers included. The training process was optimized using Adam as the optimizer with a learning rate of 0.00001, and sparse categorical cross-entropy was used as the loss function. Given the large size of 3D neuroimaging data, we have used the stochastic gradient descent algorithm to train the model. All models were trained for 20 epochs.
The multimodal integration was achieved through feature fusion, where extracted MRI features were combined with temporally processed fMRI features, enabling the models to learn complementary structural and functional information. The training performance was monitored using accuracy and loss metrics on both training and validation sets to ensure robust generalization.\\

\subsection*{Multimodal Model Architecture and Training}

We designed a multimodal DL architecture $\mathcal{M}_{\text{multi}}$ for the classification of AD integrating MRI and fMRI information. The model consists of two branches: a 3DCNN for MRI and a Time-Distributed 3DCNN followed by stacked LSTM layers for fMRI. The inputs are MRI $\mathbf{X}_\text{s} \in \mathbb{R}^{d_1 \times d_2 \times d_3 \times 1}$ and fMRI $\mathbf{X}_\text{f} \in \mathbb{R}^{T \times d_1 \times d_2 \times d_3 \times 1}$, where $d_1, d_2, d_3$ are spatial dimensions and $T$ is the number of temporal frames.

\paragraph{MRI Branch ($\mathcal{B}_{\text{MRI}}$):}  
The MRI is processed through a series of 3D convolutional and pooling layers, defined as:
\[
\mathbf{H}_1 = \sigma_{\text{ReLU}}\left(\mathcal{C}_{3D}^{(1)}(\mathbf{X}_\text{s})\right), \quad \mathbf{H}_1 \in \mathbb{R}^{\frac{d_1}{2} \times \frac{d_2}{2} \times \frac{d_3}{2} \times 32}
\]
\[
\mathbf{H}_2 = \sigma_{\text{ReLU}}\left(\mathcal{C}_{3D}^{(2)}\left(\mathcal{P}_{3D}(\mathbf{H}_1)\right)\right), \quad \mathbf{H}_2 \in \mathbb{R}^{\frac{d_1}{4} \times \frac{d_2}{4} \times \frac{d_3}{4} \times 32}
\]
\[
\mathbf{v}_\text{MRI} = \phi_{\text{dense}}^{128}(\text{Flatten}(\mathcal{P}_{3D}(\mathbf{H}_2))) \in \mathbb{R}^{128}
\]

\paragraph{fMRI Branch ($\mathcal{B}_{\text{fMRI}}$):}  
Each time frame of the fMRI volume is processed using a Time-Distributed wrapper over a 3DCNN, followed by stacked LSTM layers to capture temporal dependencies:
\[
\mathbf{H}_{\text{TD}} = \text{Time-Distributed}(\mathcal{B}_{\text{CNN}})(\mathbf{X}_\text{f}), \quad \mathbf{H}_{\text{TD}} \in \mathbb{R}^{T \times N}
\]
where $N$ is the output of the flattened CNN block. This is followed by:
\[
\mathbf{L}_1 = \mathcal{LSTM}^{(1)}_{32}(\mathbf{H}_{\text{TD}}), \quad 
\]
\[
\mathbf{L}_2 = \mathcal{LSTM}^{(2)}_{32}(\mathbf{L}_1), \quad 
\]
\[
\mathbf{v}_\text{fMRI} = \mathcal{LSTM}^{(3)}_{32}(\mathbf{L}_2) \in \mathbb{R}^{32}
\]

\paragraph{Fusion and Classification:}  
The two latent vectors $\mathbf{v}_\text{MRI} \in \mathbb{R}^{128}$ and $\mathbf{v}_\text{fMRI} \in \mathbb{R}^{32}$ are concatenated and passed through fully connected layers:
\[
\mathbf{z}_1 = \sigma_{\text{ReLU}}(\phi_{\text{dense}}^{64}([\mathbf{v}_\text{MRI}; \mathbf{v}_\text{fMRI}])), \quad \mathbf{z}_1 \in \mathbb{R}^{64}
\]
\[
\mathbf{z}_2 = \text{Dropout}_{0.1}(\mathbf{z}_1), \quad \mathbf{\hat{y}} = \text{Softmax}(\phi_{\text{dense}}^{C}(\mathbf{z}_2)), \quad \mathbf{\hat{y}} \in \mathbb{R}^{C}
\]
where $C=3$ is the number of output classes.

\paragraph{Training:}  
The model $\mathcal{M}_{\text{multi}}$ is trained using the Adam optimizer with a learning rate of $\eta = 10^{-5}$. The loss function used is sparse categorical cross-entropy:
\[
\mathcal{L}_{\text{CE}} = -\sum_{i=1}^{C} \mathbf{y}_i \log(\mathbf{\hat{y}}_i)
\]
where $\mathbf{y}$ is the true class label (sparse format) and $\mathbf{\hat{y}}$ is the predicted probability vector. The model is evaluated using accuracy and other classification metrics described in Section ~\ref{sec:ResultsD}.

\section{Results}
\label{sec:ResultsD}
In this work, to evaluate the performance of AD classification model, we utilized several standard metrics: Area Under the Receiver Operating Characteristic Curve  (AUC) \cite{ROC-AUC-Chicco}, Confusion Matrix ($\mathcal{C}_\mathrm{M}$) \cite{Confusion-Matrix}, Precision ($\mathcal{P}$), Recall ($\mathcal{R}$) \cite{Precision-Recall-F1Score}, and F1-score ($\mathcal{F}_1$) \cite{MCC-F1-score}. These are defined as follows:  
Precision: $\mathcal{P} = \frac{\mathcal{TP}}{\mathcal{TP} + \mathcal{FP}}$,  
Recall: $\mathcal{R} = \frac{\mathcal{TP}}{\mathcal{TP} + \mathcal{FN}}$,  
F1-score: $\mathcal{F}_1 = 2 \cdot \frac{\mathcal{P} \cdot \mathcal{R}}{\mathcal{P} + \mathcal{R}}$,  
where $\mathcal{TP}$, $\mathcal{FP}$, and $\mathcal{FN}$ represent true positives, false positives, and false negatives, respectively. The AUC measures the model’s ability to distinguish between classes across various threshold settings, and the $\mathcal{C}_\mathrm{M}$ provides a comprehensive view of classification outcomes.

\subsection{Without Augmentation}
\begin{figure*}[!t]
    \centering
    \includegraphics[width=0.48\textwidth]{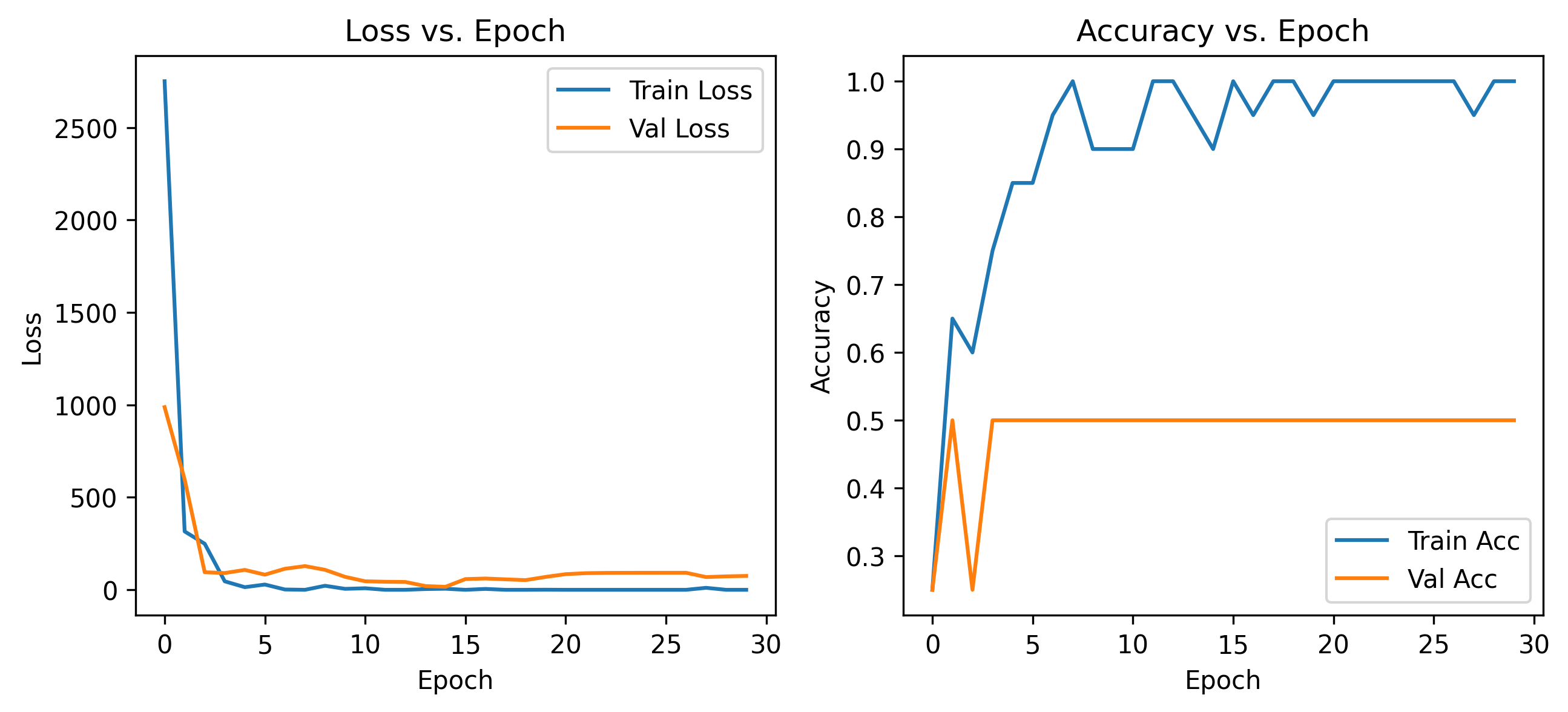} \hfill
    \includegraphics[width=0.48\textwidth]{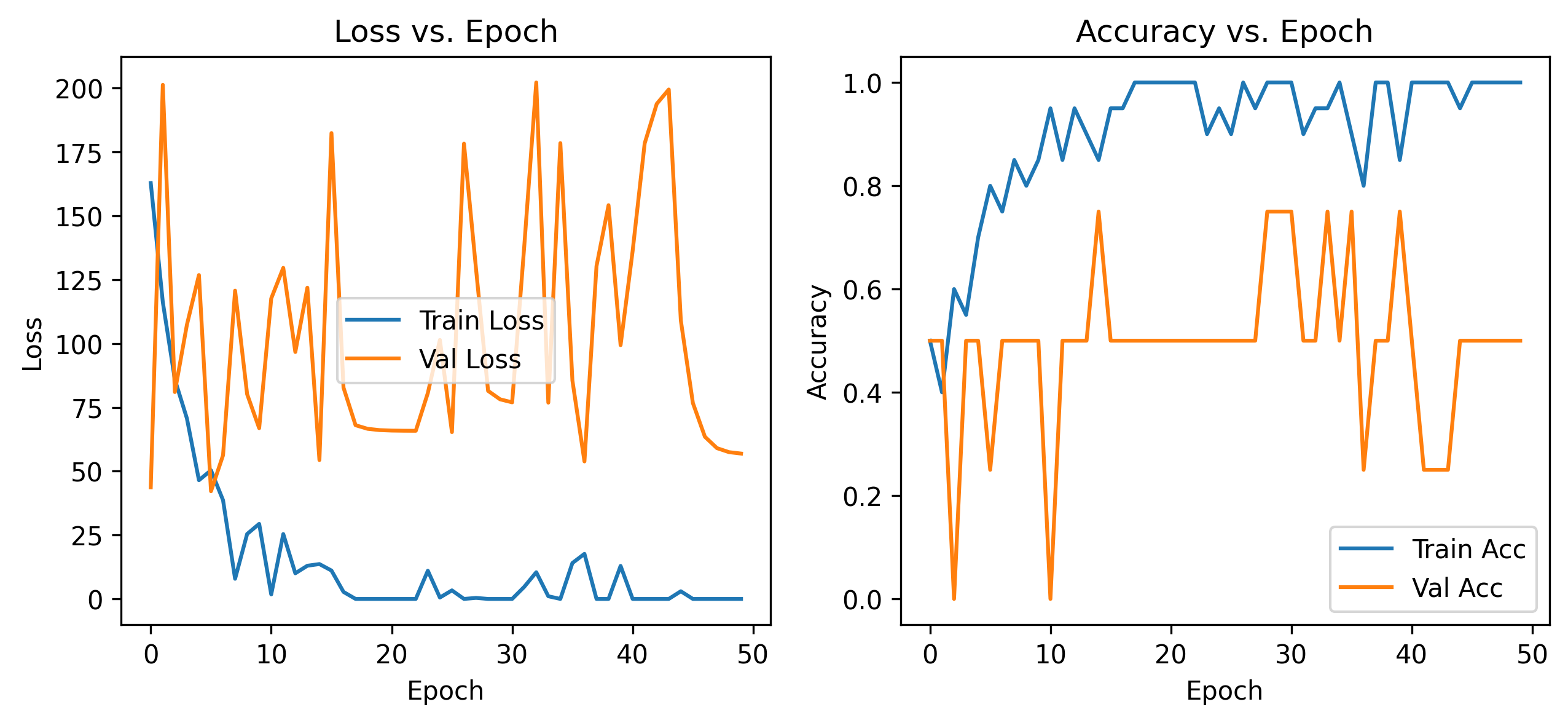}
    \caption{Learning Curves: (a) 3DCNN-LSTM (b) 3DCNN-GRU}
    \label{fig:training_acc_loss}
\end{figure*}

\begin{figure*}[t]
    \centering
    \includegraphics[width=0.48\textwidth]{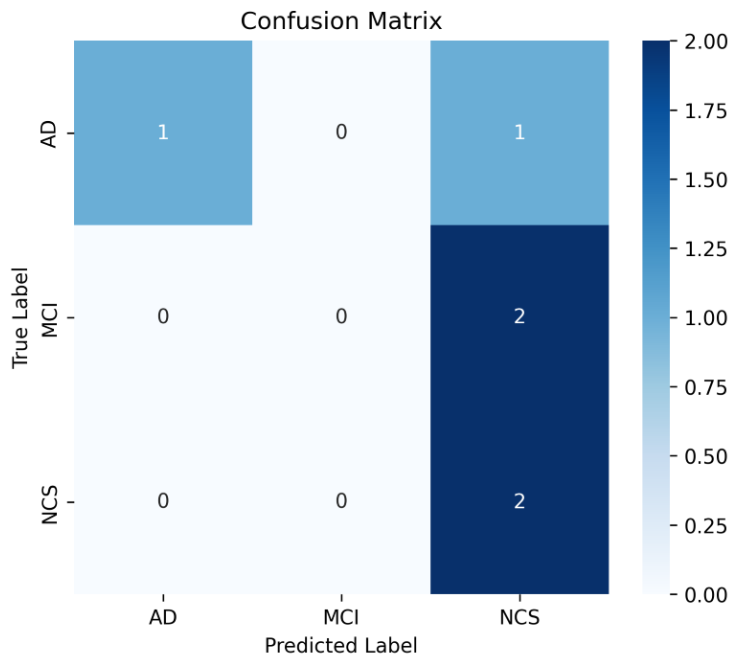} \hfill
    \includegraphics[width=0.48\textwidth]{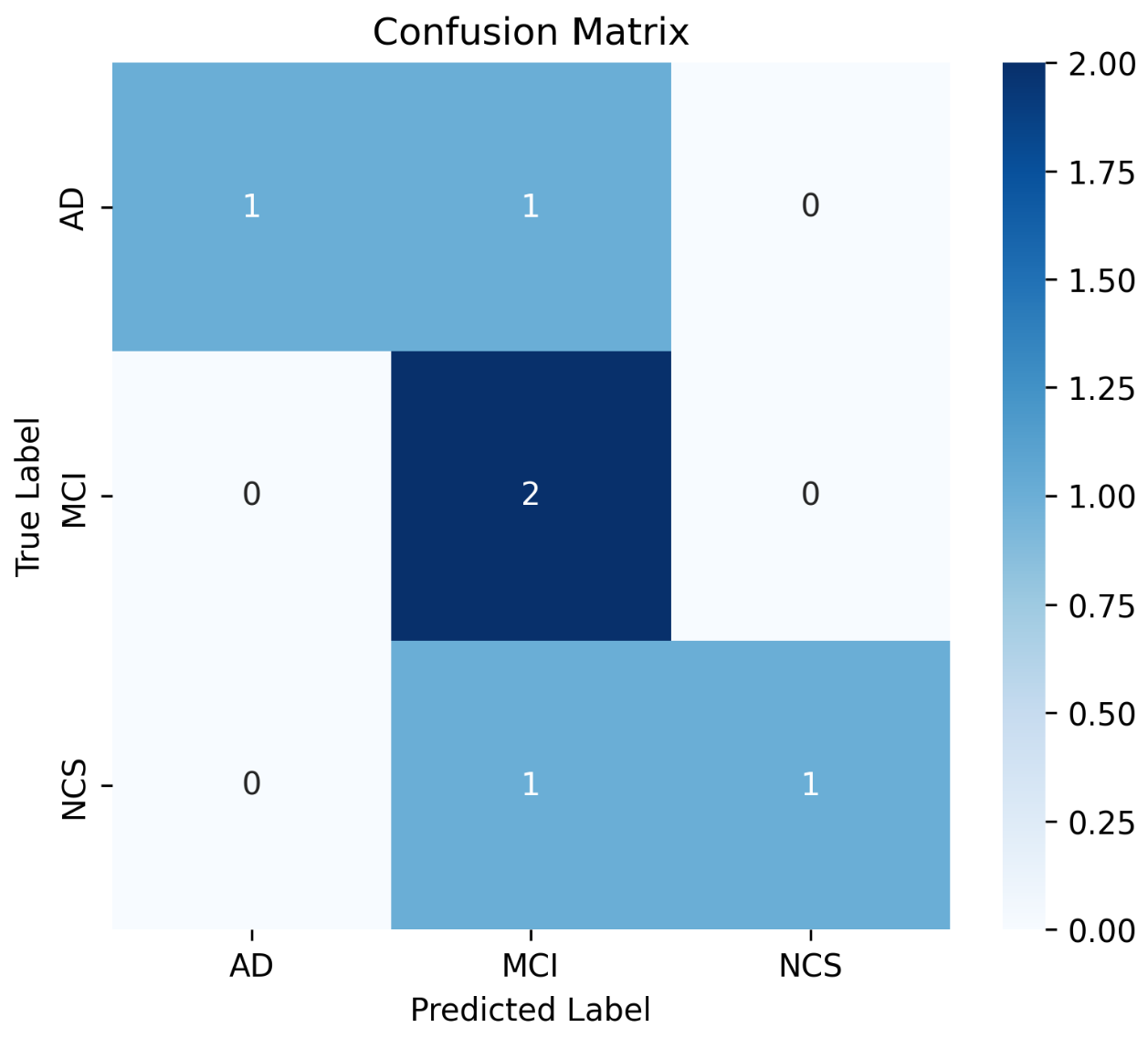}
    \centering
    \caption{CM:(a) 3DCNN-LSTM (b) 3DCNN-GRU}
    \label{fig:conf_matrices}
\end{figure*}

\begin{figure}[h!]
\centering\includegraphics[width=0.8\textwidth]{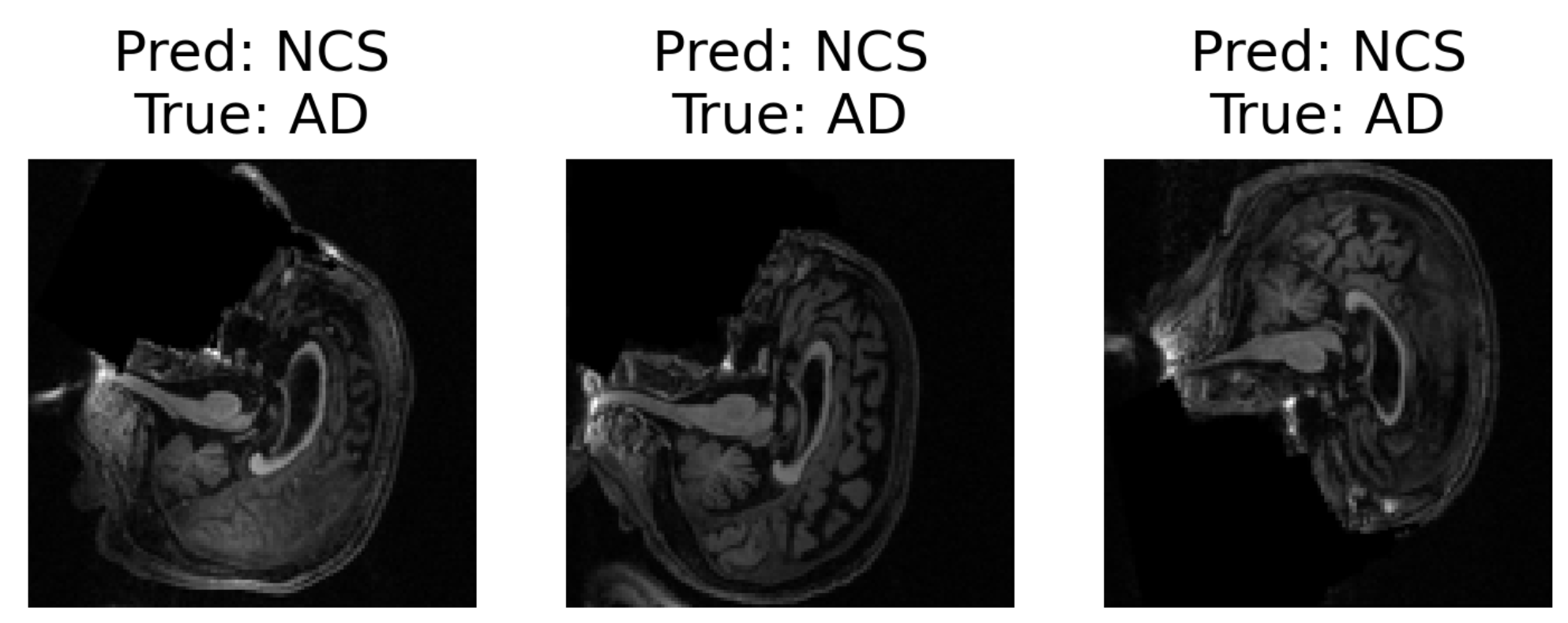}
    \caption{Examples of incorrectly classified MRI slices from the non-augmented HD dataset. The sample indices correspond to internal test set ordering and are shown only for reference.}
    \label{fig:predicted images}
\end{figure}

Without data augmentation, DL models trained on limited medical imaging data often suffer from significant drawbacks. In cases where the dataset is minuscule, the model may rely on random "lucky guesses" rather than learning meaningful patterns, leading to unreliable predictions. Additionally, the model tends to overfit to narrow, dataset-specific patterns instead of capturing generalizable and clinically relevant features. This results in poor inter-patient variability, where the model fails to generalize across different subjects, limiting its real-world applicability. Moreover, the absence of augmentation leads to unstable parameter estimates, making the model highly sensitive to minor changes in the input data. Consequently, without augmentation, DL models struggle to achieve robust and reliable performance, reducing their effectiveness in critical applications such as medical image analysis.
The performance of the proposed multimodal DL framework was first evaluated without applying any data augmentation techniques. Figures ~\ref{fig:training_acc_loss}, ~\ref{fig:conf_matrices}, and ~\ref{fig:predicted images} provide detailed insights into the training behavior, classification performance, and qualitative predictions of the model variants: 3DCNN-LSTM and 3DCNN-GRU.
Fig.~\ref{fig:training_acc_loss} shows the training and validation loss and accuracy curves for both the 3DCNN-LSTM and 3DCNN-GRU architectures.
For the 3DCNN-LSTM model (Fig.~\ref{fig:training_acc_loss}a), the training loss decreases sharply within the first 10 epochs and stabilizes at a low value, indicating effective learning. The validation loss also shows a decreasing trend but converges at a higher value compared to the training loss, suggesting potential overfitting. The training accuracy reaches nearly 100\%, while the validation accuracy plateaus around 55\%, further highlighting the generalization gap.

In contrast, the 3DCNN-GRU model (Fig.~\ref{fig:training_acc_loss}b) exhibits unstable training behavior. The training loss decreases steadily, but the validation loss fluctuates significantly throughout the epochs. This instability is mirrored in the validation accuracy, which varies greatly and fails to converge. These results suggest that the GRU-based model struggles more with learning consistent temporal-spatial features from the input data compared to the LSTM variant.

Fig.~\ref{fig:conf_matrices} presents the confusion matrices for the two model variants, providing insight into class-specific prediction performance across the three classes: AD, MCI, and NCS.

For the 3DCNN-LSTM model (Fig.~\ref{fig:conf_matrices}a), while AD and NCS are predicted with some accuracy, the model shows confusion between AD and NCS, and fails to identify MCI cases accurately. Similarly, the 3DCNN-GRU model (Fig.~\ref{fig:conf_matrices}b) misclassifies some AD samples as MCI and shows errors in distinguishing between MCI and NCS. Overall, both models perform poorly on minority classes, possibly due to class imbalance and limited data.

Qualitative Prediction Analysis from Fig.~\ref{fig:predicted images} shows example predictions where the model misclassifies AD cases as NCS. These results emphasize the challenge of learning robust discriminative features in the absence of augmentation, which might have enhanced generalization. The anatomical differences in these samples, though subtle, are critical and require more training data diversity to be effectively captured.

These findings underscore the need for improved data diversity and regularization, motivating the incorporation of data augmentation strategies in subsequent experiments.

\subsection{With Augmentation}
With data augmentation, DL models benefit from increased variability in the training data, leading to more stable and reliable performance. Classification reports tend to converge toward consistent patterns, reducing randomness in predictions and improving generalization. Although overfitting may still persist, it becomes more controlled, demonstrating consistency across key evaluation metrics. This results in meaningful scores that better reflect the model's true capability rather than being skewed by limited or biased data. Furthermore, augmentation improves statistical validity, making model performance interpretable and more applicable to real-world scenarios, particularly in complex tasks such as medical image analysis.

To evaluate the effectiveness of our data augmentation strategy and the performance of our proposed multimodal DL models, we conducted a series of experiments using 3DCNN combined with recurrent architectures. The results, presented in Figures ~\ref{fig:fmri roc_aug}, ~\ref{fig:kdata_roc}, and ~\ref{fig:kdata_cm} demonstrate improved classification performance across AD, MCI, and NCS following data augmentation.

Fig.~\ref{fig:fmri roc_aug}(a) and (b) illustrates the multi-class ROC curves for the 3DCNN-LSTM model trained on the augmented HD dataset. 
Fig. ~\ref{fig:predicted images} shows prediction results on three representative MRI slices from the test set. The visual predictions with augmentation indicate correct model predictions for various cases. For instance, the model accurately predicts AD for sample index 9 and 26, and MCI for index 11, matching the ground truth. These qualitative assessments confirm that the augmented data effectively enhances the model's ability to learn discriminative spatial features relevant to diagnosis.

\subsection{Further Parameter Tuning}

To assess the sensitivity of recurrent architectures under identical training conditions, we compared 3DCNN–LSTM and 3DCNN–GRU models on the augmented Harvard Dataverse dataset. No additional datasets or optimization procedures were introduced beyond those described earlier. The comparison focuses on classification performance rather than hyperparameter optimization.

\begin{figure*}[!t]
    \centering
    \includegraphics[width=0.48\textwidth]{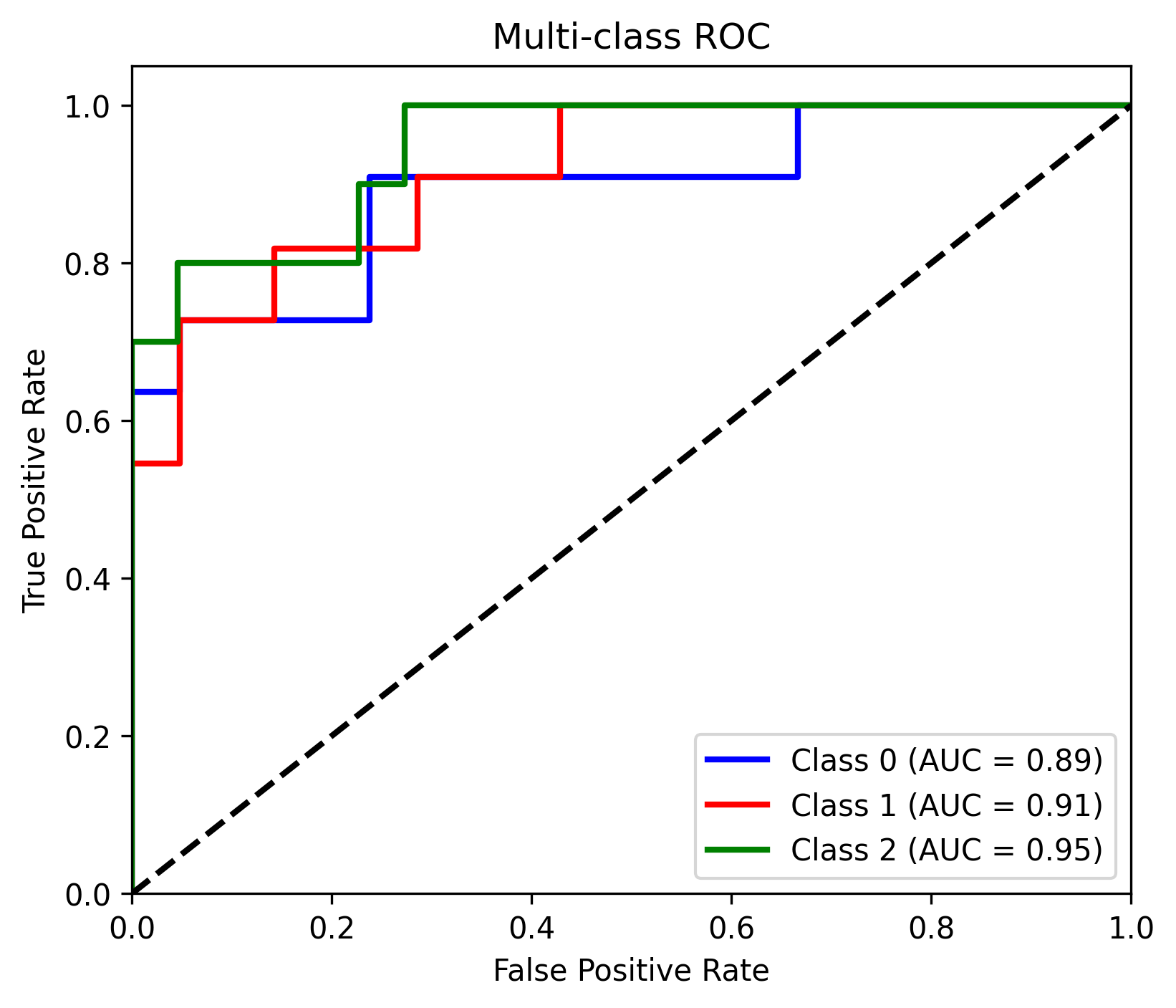} \hfill
    \includegraphics[width=0.48\textwidth]{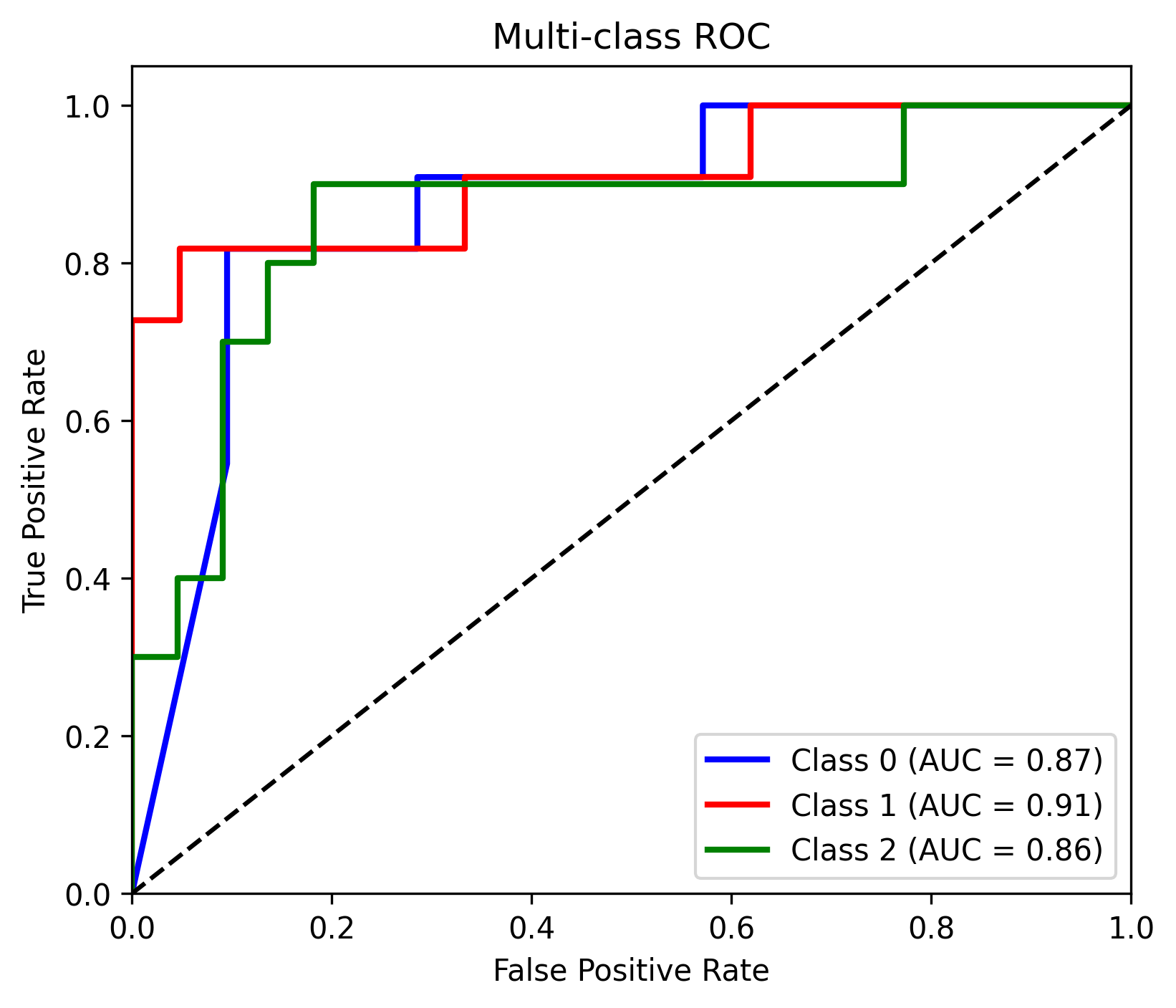}
     \caption{ROC curves for multimodal classification on the HD dataset. (a) 3DCNN–LSTM model with augmentation. (b) 3DCNN–GRU model with augmentation}
    \label{fig:fmri roc_aug}
\end{figure*}

To further optimize performance, we performed parameter tuning and evaluated alternative recurrent architectures. Fig.~\ref{fig:fmri roc_aug} compares ROC curves for two models: 3DCNN-LSTM and 3DCNN-GRU. The 3DCNN-LSTM model achieves AUC scores of 0.89, 0.91, and 0.95 for Classes 0, 1, and 2, respectively, marking a substantial improvement, especially for AD detection. Similarly, the 3DCNN-GRU model yields AUC scores of 0.87 (NCS), 0.91 (MCI), and 0.86 (AD), signifying competitive performance. These results suggest that GRU-based models, while slightly less performant for Class 2, still offer promising results with reduced computational overhead compared to LSTM.

Overall, the results substantiate the critical role of data augmentation in improving model generalization, reducing overfitting, and enhancing diagnostic accuracy in the context of neuroimaging-based AD classification. The recurrent components in our architecture effectively capture temporal dependencies in fMRI data, while convolutional layers extract relevant spatial features from both structural and functional modalities, confirming the robustness of our unified DL framework.

\subsection{Impact of Data Augmentation and Model Simplicity on Large MRI Datasets}
\begin{figure}[t]
    \centering
    \includegraphics[width=0.48\textwidth]{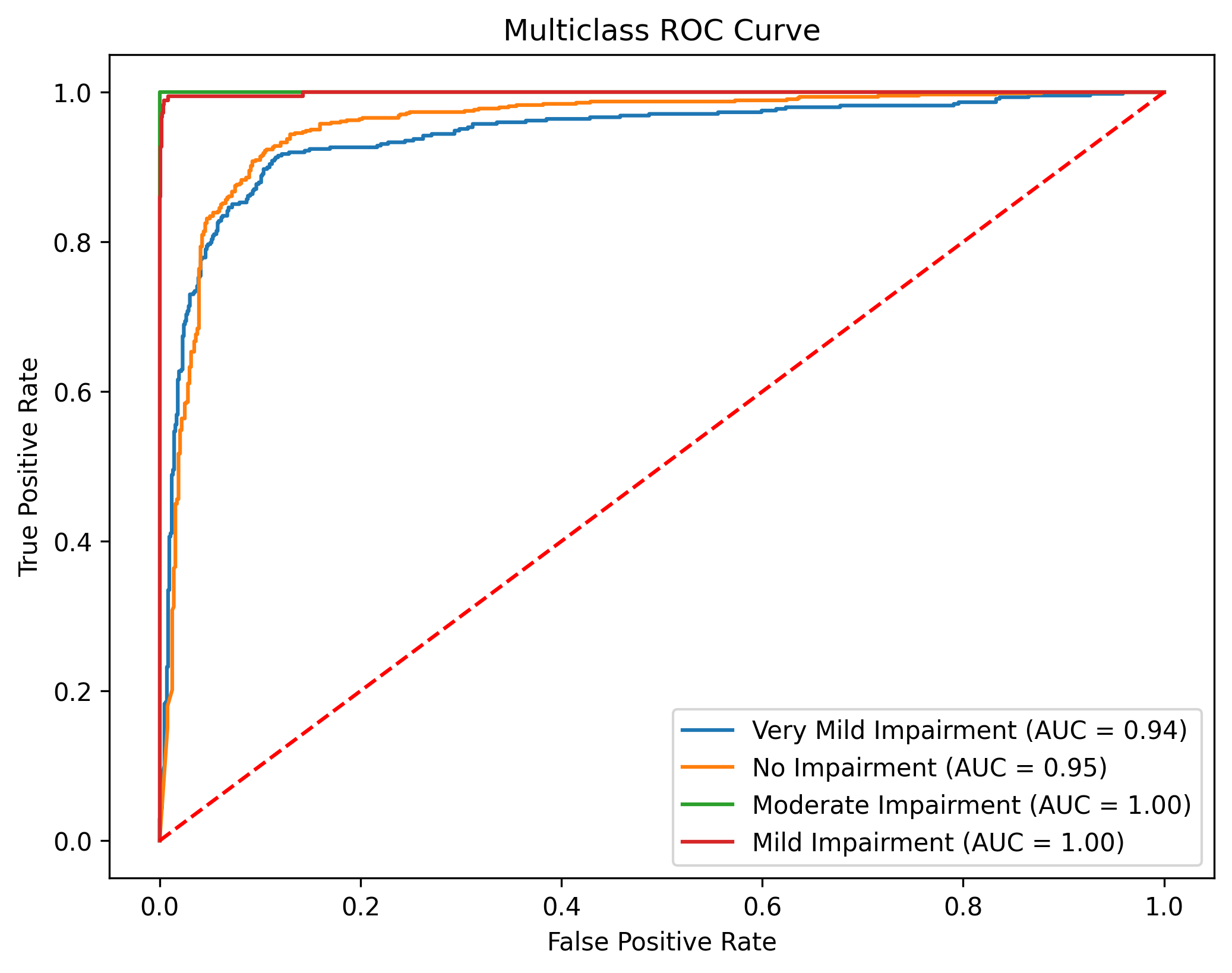} \hfill
    \includegraphics[width=0.48\textwidth]{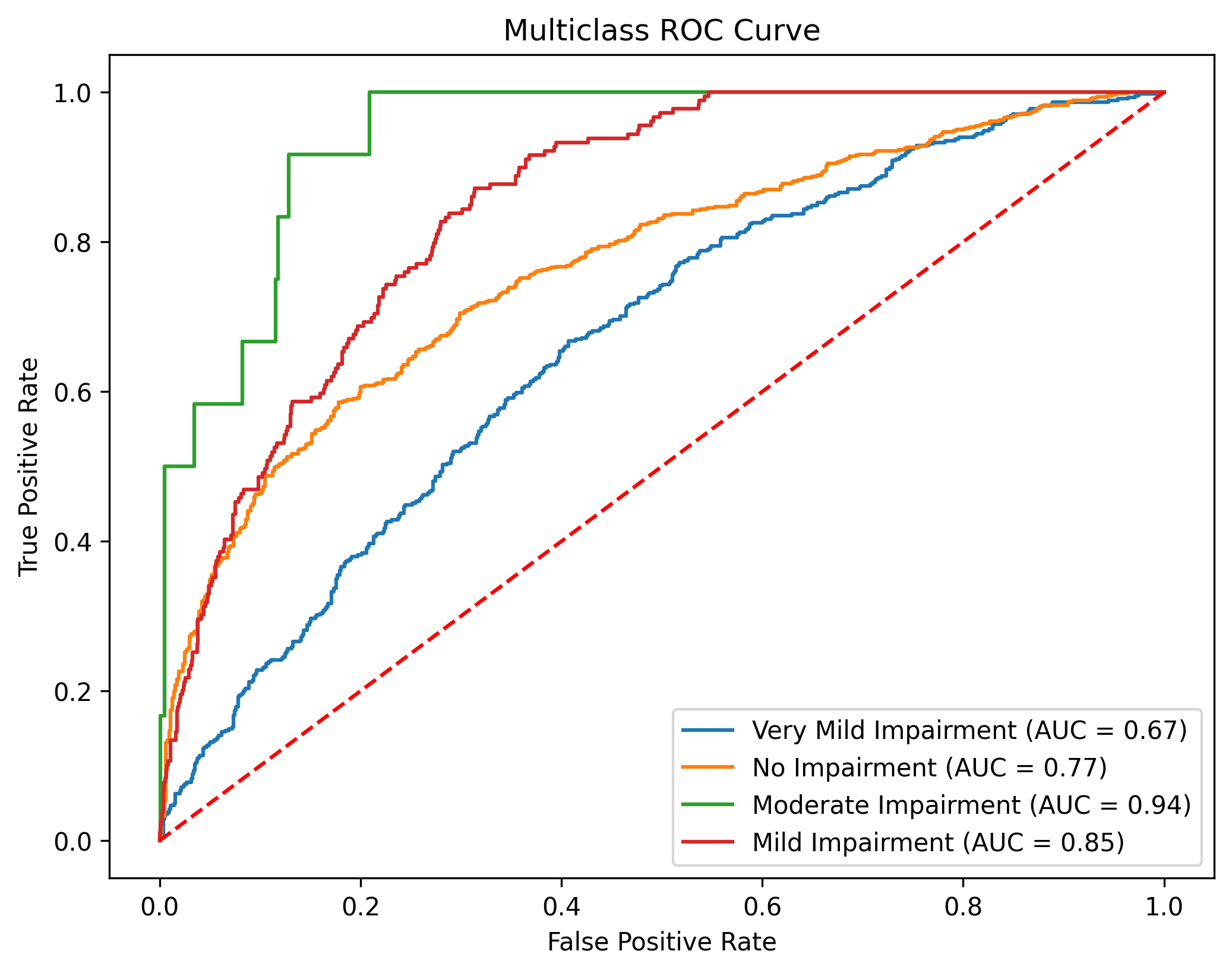}
    \caption{ROC curves: (a) Without Augmentation (Kaggle Data) (b) With Augmentation (Kaggle Data)}
    \label{fig:kdata_roc}
\end{figure}

\begin{figure}[t]
    \centering
    \includegraphics[width=0.48\textwidth]{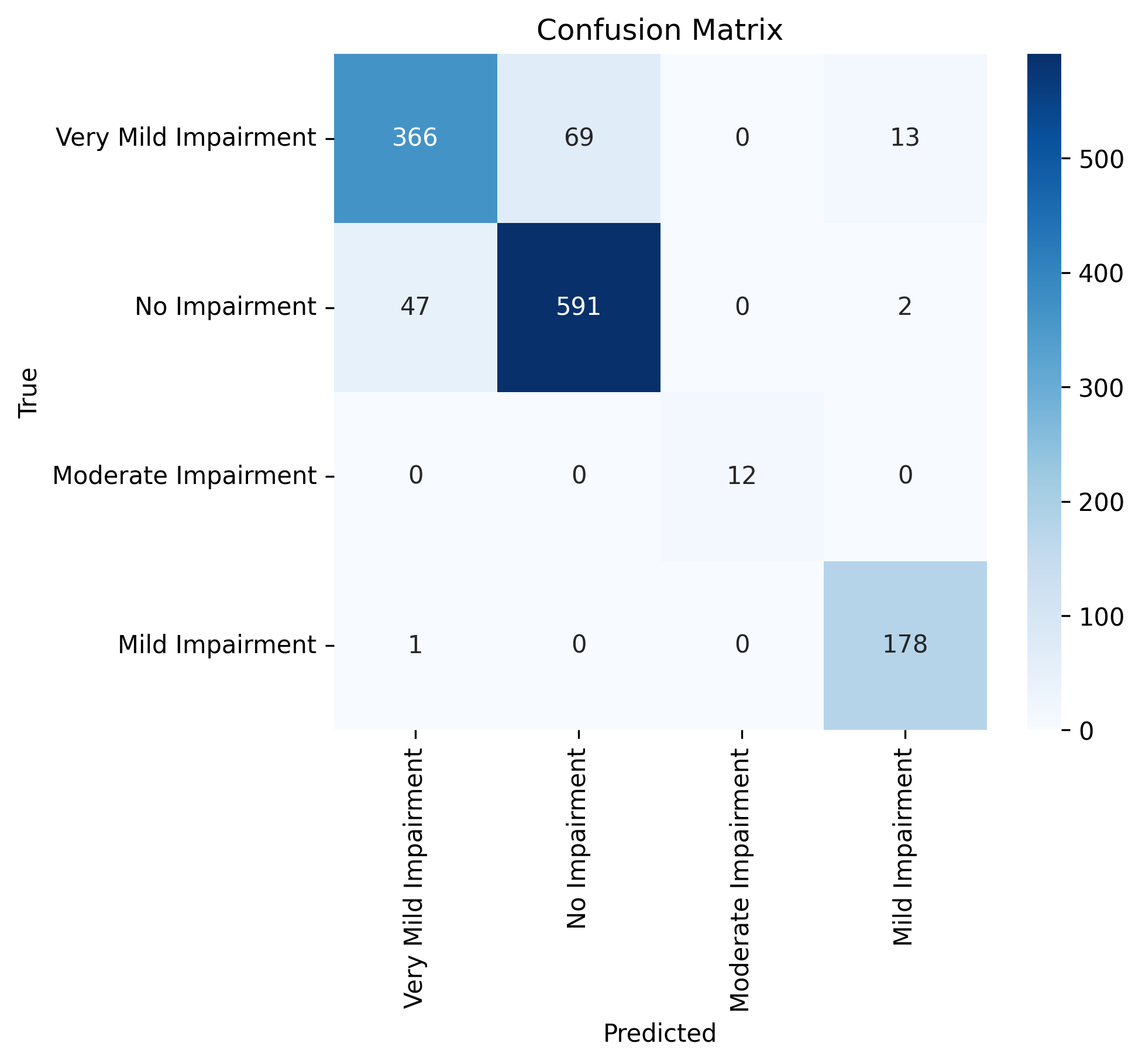} \hfill
    \includegraphics[width=0.48\textwidth]{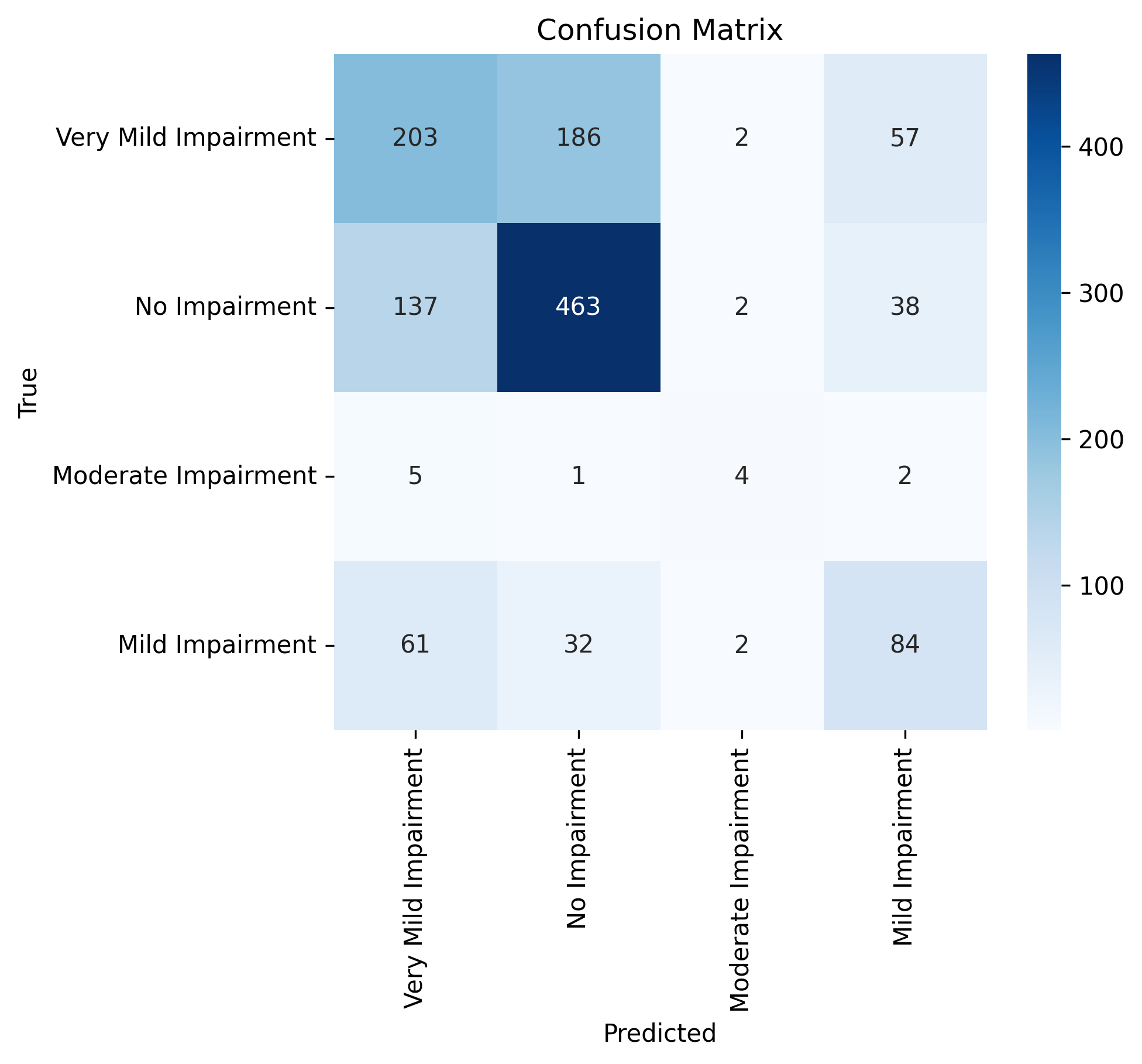}
    \caption{CM: (a) Without Augmentation (Kaggle Data) (b) With Augmentation (Kaggle Data)}
    \label{fig:kdata_cm}
\end{figure}

To evaluate the role of data augmentation in the context of large MRI datasets, a series of experiments were conducted using a simplified 2DCNN architecture. The dataset comprised T1-weighted MRI scans labeled across four cognitive impairment categories: Very Mild Impairment, No Impairment, Mild Impairment, and Moderate Impairment. As illustrated in Fig.~\ref{fig:kdata_roc} and ~\ref{fig:kdata_cm}, the results clearly show the performance contrast before and after data augmentation was applied.

Fig.~\ref{fig:kdata_roc} compares the multiclass ROC curves before and after augmentation. Without augmentation (Fig.~\ref{fig:kdata_roc}, left), the model achieves outstanding AUC values: 0.94 for Very Mild Impairment, 0.95 for No Impairment, and a perfect 1.00 for both Moderate and Mild Impairments, indicating excellent separability between classes.

Conversely, after applying augmentation (Fig.~\ref{fig:kdata_roc}, right), AUC scores drop significantly. For instance, Very Mild Impairment and No Impairment drop to 0.67 and 0.77, respectively. The performance drop indicates that augmentation may have introduced artifacts or distortions that negatively impacted the model’s ability to extract meaningful and discriminative features.
Fig.~\ref{fig:kdata_cm} (left) shows the confusion matrix for the model trained without augmentation. The model performs strongly across most classes, particularly for No Impairment and Very Mild Impairment, achieving high true positive rates with limited misclassifications. For example, the model correctly identifies 591 out of 640 samples labeled as No Impairment, misclassifying only a small fraction.

However, after introducing data augmentation (Fig.~\ref{fig:kdata_cm}, right), classification performance degrades noticeably. The model becomes less confident in distinguishing between classes, particularly between Very mild impairment and no impairment, as well as between mild impairment and other classes. 
The observed performance degradation suggests that the applied augmentation strategy may have altered discriminative image characteristics in this dataset. In addition to dataset size, factors such as image preprocessing, class heterogeneity, and acquisition variability may also influence augmentation effectiveness.

These results support the hypothesis that with a sufficiently large and diverse dataset, as is the case here, the inclusion of synthetic augmentations may reduce generalization capability rather than enhance it. The 2DCNN, despite its simplicity, is capable of learning robust spatial features when presented with large real data. 
Introducing complex architectures or augmenting the training data without justification can disrupt the learning process, particularly when the original dataset already contains sufficient variability to capture the underlying patterns.

Thus, this experiment demonstrates that in large-scale MRI datasets, simplicity in both model design and data preprocessing yields the most stable and reliable outcomes. Data augmentation, when not carefully curated, may inject harmful variance that hinders rather than helps.

\section{Discussion}
\label{sec:Dis}
The present study introduces a comprehensive multimodal DL framework integrating MRI and fMRI to advance the classification of AD, MCI, and NCS. By utilizing both spatial and temporal features, the proposed framework enables a c analysis of neurodegenerative classification. In this section, we compare and critically analyze the experimental design, model architectures, augmentation strategies, and performance outcomes. The fusion of MRI and fMRI enabled the model to jointly learn static anatomical and dynamic functional patterns, significantly enhancing its discriminative capability. The 3DCNN-based MRI encoder effectively captured high-resolution volumetric features, while the Time-Distributed CNN-LSTM block extracted time-dependent patterns from fMRI sequences. This spatial-temporal integration reflects the pathophysiological nature of AD, which occurs through both cortical atrophy and disrupted brain activity.

The experimental results demonstrated that the combination of CNN for spatial encoding and LSTM for temporal modeling leads to superior performance in capturing the complex neurodynamics compared to CNN or RNN alone. This is aligned with earlier studies that emphasize the importance of modality-aware architectures in neuroimaging-based diagnostics.

A particularly contribution of this work lies in the design of a domain-specific data augmentation pipeline for both MRI and fMRI modalities. The augmentation techniques used are spatial rotation, intensity scaling, temporal distortion, and noise injection.
Augmentation improved performance for models trained on the small paired MRI–fMRI dataset, whereas no improvement was observed for the large-scale Kaggle MRI dataset.

After applying augmentation, the 3DCNN-LSTM model achieved AUC scores of 0.92 (NCS), 0.86 (MCI), and 0.76 (AD), compared to lower and inconsistent performance in the non-augmented case. Additionally, the confusion matrix showed balanced improvements across classes. These findings confirm that augmentation not only addresses class imbalance but also improves model robustness in clinical neuroimaging applications. The study evaluated a suite of architectures, including 3DCNN-LSTM, 3DCNN-GRU, standalone 3DLSTM and 3DGRU, and a 2DCNN baseline. 
the 3DCNN–LSTM achieved the highest AUC values on the augmented multimodal dataset, while GRU-based variants showed comparable performance with lower computational complexity.
Interestingly, in scenarios with larger datasets (e.g., MRI-only experiment with over 8000 samples), the simple 2DCNN outperformed its deeper counterparts when no augmentation was applied. In this case, augmentation degraded performance, indicating that in 
datasets with a large number of samples and substantial inherent variability, artificial perturbations may introduce learning noise. Therefore, model complexity and augmentation strategies must be adapted to dataset size and diversity.

Table~\ref{tab:model_comparison} summarizes the comparative insights of all model architectures and training strategies, detailing key hyperparameters, input modalities, and evaluation metrics such as accuracy and AUC.

\begin{table}[t]
\centering
\caption{Comparison of models, input modalities, and performance}
\label{tab:model_comparison}
\renewcommand{\arraystretch}{1.15}
\begin{tabularx}{\textwidth}{l c c c c c}
\toprule
\textbf{Model} &
\textbf{Modality} &
\textbf{Temporal} &
\textbf{Augment.} &
\textbf{Data Size} &
\textbf{AUC (AD/MCI/NCS)} \\
\midrule
3DCNN--LSTM & MRI + fMRI & LSTM & Yes & 319 & 0.76 / 0.86 / 0.92 \\
3DCNN--GRU  & MRI + fMRI & GRU  & Yes & 319 & 0.86 / 0.91 / 0.87 \\
3DLSTM     & fMRI only  & Yes  & No  & 29  & Low stability \\
3DGRU      & fMRI only  & Yes  & No  & 29  & Fluctuating \\
2DCNN      & MRI only   & No   & No  & 8192+ & 0.94+ \\
2DCNN      & MRI only   & No   & Yes & 8192+ & 0.67 drop \\
\bottomrule
\end{tabularx}
\end{table}

Beyond data augmentation, overfitting was partially controlled through architectural simplicity, limited training epochs, and validation-based performance monitoring. No additional regularization strategies were explored. 

\section{Conclusion}
\label{sec:Con}
The results demonstrate the feasibility of multimodal spatial–temporal learning within the evaluated datasets. Given the limited availability of multimodal data, data augmentation emerges as a crucial technique to enhance model generalization. However, despite augmentation efforts, the number of augmented samples remains relatively small compared to the vast variability in real-world scenarios. Nonetheless, this approach marks a significant step towards bridging the gap between static and dynamic neuroimaging data. Future research should focus on improving augmentation strategies, leveraging self-supervised learning and more sophisticated generative models to strengthen the generalizability of multimodal DL frameworks.
\paragraph*{Limitations and Future Work:} Despite promising results, several limitations must be acknowledged. The dataset size remains small, even after augmentation, which could affect the reproducibility of findings. Furthermore, differences in imaging protocols and inter-subject variability introduce hidden biases. Future work should aim to:

\begin{itemize}
    \item Integrate more diverse and multi-institutional datasets.
    \item Explore transformer-based encoders for modeling long-range dependencies.
    \item Implement attention-guided fusion and contrastive learning for better cross-modal alignment.
    \item Develop generative models to synthesize realistic paired MRI/fMRI data.
    \item Investigate domain adaptation and self-supervised learning for enhanced generalization.
\end{itemize}

\section*{Acknowledgment}
Z.M. is grateful to the Bundesministerium für Bildung und Forschung (BMBF, Federal Ministry of Education and Research) for funding through project OIDLITDSM, No. 01IS24061.

\bibliographystyle{unsrt}
\bibliography{references}   
\end{document}